\documentclass{article}

\usepackage{microtype}
\usepackage{graphicx}
\usepackage{subfigure}
\usepackage{booktabs} 
\usepackage{float} 
\usepackage{algorithmic}
\usepackage{algorithm}

\usepackage{hyperref}

\usepackage[accepted]{icml2025}

\usepackage{amsmath}
\usepackage{amssymb}
\usepackage{mathtools}
\usepackage{amsthm}

\usepackage[capitalize,noabbrev]{cleveref}

\theoremstyle{plain}

\theoremstyle{definition}

\theoremstyle{remark}

\usepackage[textsize=tiny]{todonotes}

\icmltitlerunning{Learned Weight Distance Function for Nearest Neighbor Search with Multiple Filters}

\begin{document}

\twocolumn[
\icmltitle{Learning Filter-Aware Distance Metrics for Nearest Neighbor Search with Multiple Filters}



\icmlsetsymbol{}{}

\begin{icmlauthorlist}
\icmlauthor{Ananya Sutradhar}{mri}
\icmlauthor{Suryansh Gupta}{mri}
\icmlauthor{Ravishankar Krishnaswamy}{mri}
\icmlauthor{Haiyang Xu}{ms}
\icmlauthor{Aseem Rastogi}{mri}
\icmlauthor{Gopal Srinivasa}{mri}

\end{icmlauthorlist}

\icmlaffiliation{mri}{Microsoft Research India}
\icmlaffiliation{ms}{Microsoft Corporation}

\icmlcorrespondingauthor{Suryansh Gupta}{suryangupta@microsoft.com}
\icmlcorrespondingauthor{Gopal Srinivasa}{gopalsr@microsoft.com}
\icmlcorrespondingauthor{Ananya Sutradhar}{t-asutradhar@microsoft.com}

\icmlkeywords{ICML, ANN, DiskANN, Filtered DiskANN, Linear Programming, RAG, Vector Search}

\vskip 0.3in

]



\printAffiliationsAndNotice{}  


\begin{abstract}
Filtered Approximate Nearest Neighbor (ANN) search retrieves the closest vectors for a query vector from a dataset. It enforces that a specified set of discrete labels $S$ for the query must be included in the labels of each retrieved vector. Existing graph-based methods typically incorporate filter awareness by assigning fixed penalties or prioritizing nodes based on filter satisfaction.
However, since these methods use fixed, data independent penalties, they often fail to generalize across datasets with diverse label and vector distributions.

In this work, we propose a principled alternative that learns the optimal trade-off between vector distance and filter match directly from the data, rather than relying on fixed penalties. We formulate this as a constrained linear optimization problem, deriving weights that better reflect the underlying filter distribution and more effectively address the filtered ANN search problem. These learned weights guide both the search process and index construction, leading to graph structures that more effectively capture the underlying filter distribution and filter semantics.

Our experiments demonstrate that adapting the distance function to the data significantly improves accuracy by 5-10\% over fixed-penalty methods, providing a more flexible and generalizable framework for the filtered ANN search problem.
\end{abstract}

\section{Introduction}

Embedding based representations have become the cornerstone of modern machine learning systems, particularly in applications involving retrieval, search and recommendations. With the recent surge in Retrieval Augmented Generation (RAG) and other retrieval based AI systems, \emph{vector search} has emerged as one of the most critical primitives in modern information retrieval and machine learning infrastructure.

At the core of vector search lies the problem of \emph{Approximate Nearest Neighbor} (ANN) search, which aims to retrieve the top-\(k\) closest vectors to a given query vector under some distance metric. ANN search is a well studied problem and has seen widespread adoption due to its efficiency and scalability in high dimensional spaces.

However, traditional ANN search overlooks the complexities of modern applications, which often involve structured metadata or constraints. In real world scenarios such as:
\begin{itemize}
    \item \textbf{E-commerce}, buyers filter products based on attributes like brand, color, or size.
    \item \textbf{Semantic image retrieval}, users may want images from specific locations or captured under certain conditions.
    \item \textbf{Document retrieval}, users may constrain results to a certain topic, language, or publication source.
\end{itemize}

These scenarios demand not only similarity in the embedding space, but also compliance with symbolic filters. 
To address this, we consider the \textbf{Filtered-ANN} problem, a generalization of ANN search with additional filtering constraints. A prominent instantiation is the \textbf{MultiFilterANN} problem: given a dataset \( X \) of \( N \) high dimensional vectors, where each vector \( v \in X \) is annotated with a set of labels \( S_v \subseteq [m] \), the goal is to retrieve the top-\(k\) (approximately) nearest vectors to a query vector \( q \) with label set \( S_q \subseteq [m] \), subject to the constraint that \( S_q \subseteq S_v \) for each returned vector \( v \). This corresponds to an \textbf{AND-style} filtering requirement, where a candidate must match \emph{all} the filters in the query.

Existing systems typically handle filtering through strict post-processing or hard-constraint enforcement. The former requires significant over-provisioning to ensure that relevant items are not discarded, while the latter often leads to broken search paths that terminate at poor local optima. To address these limitations, the MultiFilterANN paper \cite{multifilterann2025} introduced a \emph{penalty distance function}, where the distance between a query and a candidate is defined as a linear combination of its vector distance and the extent of label mismatch. 
However, fixed penalty scoring heuristics fail to adapt to the varying importance of filters across different queries or datasets. 

In this paper, we propose a \textbf{data driven distance function} that jointly models vector similarity and filter match through a weighted combination of distance and filter mismatch. By formulating the problem as a constrained optimization, we derive scoring weights that better reflect the statistical patterns and semantics of both vectors and filters in the dataset. This allows search to dynamically adjust the relative importance of satisfying a filter versus being close in embedding space, resulting in a more flexible and generalizable framework for filtered ANN search.

%


Once the distance function is learned, we can seamlessly apply existing graph-based retrieval algorithms such as DiskANN as black-box subroutines to retrieve the top$K$ neighbors with respect to this modified distance. This eliminates the need for substantial overprovisioning typically required by post-processing techniques. Since the learned distance function incorporates filter awareness directly into the search process, our method achieves high accuracy with far fewer candidate evaluations, reducing search latency without sacrificing quality.

\paragraph{Contributions}
\begin{itemize}

    \item \textbf{Learned Distance Function:} We introduce a method to learn the optimal trade off between vector similarity and filter match, enabling a more nuanced and effective ranking of candidates that smoothly favors those satisfying the filter, rather than enforcing a strict hard constraint.

    \item \textbf{Integrated Index Construction:} We incorporate the learned distance function directly into the index building phase, allowing the index to prioritize connections between vectors that share more labels while maintaining vector similarity. This results in a graph structure better aligned with filtered ANN search, improving both accuracy and retrieval efficiency.


    \item \textbf{Empirical Validation:} We demonstrate significant performance gains over traditional fixed penalty baselines across multiple datasets, showcasing the adaptability and effectiveness of our method.
\end{itemize}

\section{Background and Related Work}

ANN Search has been studied extensively over the past few decades \cite{malkov2016efficient, datar2004lsh, huang2015queryaware, annoy2015, diskann2019}, with research focusing on various dimensions such as improving recall, scale and cost efficiency \cite{babenko2012inverted,baranchuk2018revisiting,malkov2016efficient}, real time updates \cite{singh2021freshdiskann}, distributed indexing \cite{sundaram2013streaming}. Several benchmark efforts \cite{aumuller2020ann, simhadri2022neurips} have also helped evaluate the practical trade offs among these methods.

More recently, there has been growing interest in filtered ANN Search, where queries include structured filters (e.g., metadata conditions) in addition to a vector query. With filtering becoming a standard requirement in ANN applications, many start-ups including Milvus \cite{milvus2022hybrid}, Pinecone \cite{pinecone2024hybrid}, Vearch \cite{vearch2022search}, Vespa \cite{vespa2022attributes}, and Weaviate \cite{weaviate_filtered_vector_search} —now provide ANN-as-a-service platforms featuring various degrees of filtering support. 

Several recent works aim to address MultiFilterANN explicitly. CAPS \cite{gupta2023caps} combines subset query data structures with ANN indices, but its performance degrades as the number of labels grows. SERF \cite{gupta2023caps} modifies the graph construction phase to  support range filters (e.g., time intervals), but its applicability is limited to single filters. ACORN \cite{patel2024acorn} encodes filters using low-rank projections and product quantization, but struggles to scale with complex Boolean predicates. IVF2 \cite{landrum2024parlayann} is the current state-of-the-art open-source solution for MultiFilterANN, but its clustering-only architecture can degrade in scenarios where predicates have poor alignment with clusters. Recent methods handle different filter types more effectively. Filtered-DiskANN \cite{filtereddiskann2023} targets OR-style filters, building graph indices using both vector and label information for efficient, high-recall retrieval. But this does not support AND-style filters. Another work \cite{multifilterann2025} focuses on AND-style filters, developing provable graph algorithms and using penalty to flexibly handle multiple filters.

\section{Problem Setup}
Here we formally define the Filtered-ANN search problem and also our approach of learning weights to solve it. To do so, we first provide some basic definitions.

Let $X = \{v_i\}_{i=1}^N$ denote the set of $N$ data vectors, where each $v_i  \in \mathbb{R}^d$. For each $i \in [N]$, we let $S_i \subseteq  [m]$ represent the set of labels associated with the data vector $v_i$. We use $q \in \mathbb{R}^d$ to denote the query vector and use $S_q \subseteq [m]$ to denote the labels associated with it. The distance between a query vector and a data vector is denoted by $d(q, v)$ (e.g., Euclidean distance, Cosine distance). Lastly, we define $m(q,v)$
 as a label match score, measuring the degree to which the data vector $v$ satisfies the filter constraint imposed by the query:
\[
    m(q, v) =  \frac{|S_q \cap S_v|}{|S_q|}
\]

The main goal of the Filtered-ANN search problem is to retrieve data vectors $v$ that are close to $q$ with respect to $d(q, v)$ while satisfying the filter constraint, that is, $S_q \subseteq S_v$.

To solve this Filtered-ANN search problem, as discussed in the Introduction, we propose a weighted distance function that aims to capture the optimal trade-off between the distance function and filter satisfiability. This distance function is formally defined as follows.
\[
D(q, v) = d(q, v) + w_m \cdot (1 - m(q, v)),
\]
where:
\begin{itemize}
    \item \( w_m \geq 0 \) controls the penalty applied to vectors that violate the filter constraint.
\end{itemize}


we use $(1 - m(q, v))$ to penalize candidates that fail to satisfy the predicate, converting the match score into a mismatch penalty.
We seek to learn the weight \( w_m \) such that, among all ground truth neighbors of a query:
\begin{align*}
D(q, v_1) &< D(q, v_2) \\
\text{where} \quad m(q, v_1) &= 1 \quad \text{and} \quad m(q, v_2) < 1
\end{align*}


By solving for such weights, we aim to construct a distance function that balances distance and filter match in a data-driven way, improving retrieval quality.

\section{Learning the Distance function}

Given a set of queries \( Q = 1, \ldots, q \), each associated with candidate vectors \( v \) characterized by distances \( d(q,v) \) and filter match indicators \( m(q,v) \), our goal is to learn weight \( w_m \) that combine these features into a distance function:
\[
D(q, v) = d(q,v) + w_m \cdot (1 - m(q,v)),
\]
which ranks vectors such that those satisfying the filter (\( m(q,v) = 1 \)) score better than those that do not.

\subsection{Ground-Truth Preference Pairs}

To learn the weight $w_m$, we start by constructing ground truth preference pairs for each query $q$ as follows:

\begin{itemize}
    \item For each query $q$, we perform an exact full scan retrieval over the dataset to obtain some top-$k$ ground truth nearest neighbors $\mathcal{N}_q \subseteq X$. During this process, we calculate both the vector distance $d(q, v)$ and the filter match indicator $m(q, v)$ for every candidate vector $v$.

    
    \item We define the set of positive examples as $\text{Pos}_q = \{i \mid v_i \in \mathcal{N}_q \text{ and } m(q, v_i) = 1\}$, i.e., vectors satisfying the filter constraint.
    \item For each $i \in \text{Pos}_q$, the set of negative examples is defined as $\text{Neg}_q(i) = \{j \mid m(q, v_j) < 1 \text{ and } d(q, v_j) < d(q, v_i)\}$.
\end{itemize}

\subsection{Linear Programming Formulation}

We formulate the learning problem as a Linear Program with the following components:

\paragraph{Variables:}
\begin{itemize}
    \item \( w_m \geq  0 \), weight for filter mismatch penalty.
    \item \( s_{q,i,j} \geq 0 \), slack variables to allow for soft violations of ranking constraints.
\end{itemize}

\paragraph{Objective:}

We minimize a combination of the filter mismatch weight $w_m$ and the total slack violations over all ranking constraints:


\[
\min \left(w_m + \alpha \cdot \frac{1}{|\mathcal{S}|} \sum_{(q, i, j) \in \mathcal{S}} s_{q,i,j}\right)
\]

where \(\mathcal{S} = \{(q, i, j) \mid i \in \text{Pos}_q,\, j \in \text{Neg}_q(i)\}\) is the set of all triplets defining the ranking constraints.

The parameter \(\alpha\) controls the trade off between minimizing the filter mismatch penalty \(w_m\) and minimizing the average slack violation. This helps the model find a good trade off between following the vector distances and allowing some flexibility in filter matching. We can use a grid search to choose the best value of \(\alpha\).


\paragraph{Constraints:}

For all \( q \), and all pairs \( (i,j) \) with \( i \in \text{Pos}_q \) and \( j \in \text{Neg}_q(i) \), enforce:
\begin{align}
d(q,v_i) + w_m \cdot (1 - m(q,v_i)) + \varepsilon \\
\leq d(q,v_j) + w_m \cdot (1 - m(q,v_j)) \nonumber
&\quad + s_{q,i,j},
\end{align}

where \( \varepsilon > 0 \) is a margin parameter to ensure robustness. In our experimental setting, we use \( \varepsilon = 0.01 \).

\paragraph{Goal:}
This LP formulation seeks weights that maximize the margin by which positive, filter satisfying vectors outrank negative vectors. We aim to minimize \( w_m \) because a smaller value encourages the search process to treat filter constraints more flexibly. In the extreme case, \( w_m = \infty \) enforces filters as hard constraints, disallowing any violations and potentially hurting recall. Slack variables prevent infeasibility in cases where perfect ranking is impossible.  By solving this program, we learn a data driven trade off between vector similarity and filter matching.







\section{Penalty Aware Index Construction}

We leverage the learned distance function to guide the construction and search within the nearest neighbor graph. Specifically, the weight \( w_m \) influence how edges are added and prioritized, balancing vector proximity and filter match penalties.

\subsection{Distance Function}


The combined distance between two vectors \( v_1 \) and \( v_2 \) is defined as:
\[
D(v_1, v_2) = d(v_1, v_2) + w_m \cdot (1 - m(S_{v_1}, S_{v_2})),
\]
where \( d(v_1, v_2) \) is the standard vector distance 
(e.g., cosine distance), and \((1 - m(S_{v_1}, S_{v_2}))\) measures the dissimilarity between the label sets of \( v_1 \) and \( v_2 \).

The asymmetric Jaccard distance between the label sets \( S_{v_1} \) and \( S_{v_2} \) is defined as:
\[
m(S_{v_1}, S_{v_2}) =  \frac{|S_{v_1}\cap S_{v_2}|}{|S_{v_1} |}.
\]

We use $D(v_1, v_2)$  as the distance metric during the construction of the graph-based index over the dataset. 





\subsection{Comparison with Prior Methods}

Most existing graph-based filtered search methods incorporate filter constraints using heuristic rules during search but not during index construction. A common strategy is to ignore filter labels entirely while building the index, and then apply hard constraints, such as discarding neighbors with insufficient label overlap, at query time \cite{multifilterann2025}.

In contrast, our approach integrates filter awareness directly into the index construction phase. We learn a soft distance function that linearly combines vector distance with a penalty for filter mismatch, weighted by a learned parameter \( w_m \). This enables the index to preserve edges to candidates with mild filter mismatches if their vector proximity is sufficiently high, producing a more flexible and effective graph structure. By avoiding fixed heuristics and instead adapting to data characteristics, our method constructs filter aware indices that support higher quality retrieval.

\section{Implementation Overview}


In this section, we describe the learning of a data-driven filter-aware distance function, its integration into the construction of a graph index, and the corresponding search and planning strategies. We also outline our approach to query planning, which selects an appropriate search strategy for each query.

\paragraph{Learning Filter-Aware Distance Function:}
Prior to index construction, we split the query set into two disjoint subsets: a training set used for learning, and a test set for evaluation. For each training query, we retrieve its ground truth neighbors via exhaustive unfiltered search and compute the filter match score for each result. We formulate a linear program based on these statistics that models pairwise ranking constraints between satisfying and non satisfying vectors, as described in Section 4. We solve this LP using the \texttt{PuLP} solver to obtain optimal weights for combining vector distance and filter mismatch in the distance function.

\paragraph{Building the Graph Index:}  
We construct our graph based index using FilteredDiskANN~\cite{filtereddiskann2023}, applying a modified greedy search procedure we refer to as \textit{WeightedGreedySearch}. This search integrates our learned distance function during edge selection, encouraging connections between vectors that are not only spatially close but also exhibit high filter similarity. The complete construction algorithm is provided in Algorithm~\ref{alg:build}.

\paragraph{Searching the Graph Index:}  
At query time, we utilize a \textit{penalized search} approach guided by our learned distance function. This strategy favors candidates that minimize the joint objective over vector distance and filter mismatch. Full algorithm is provided in Algorithm~\ref{alg:search}.

\paragraph{Query Planning:}  
Inspired by prior work~\cite{multifilterann2025}, we employ a simple planning mechanism that routes queries to either the graph based search or an exact search based on filter selectivity. For queries with highly selective label sets (e.g., only one matching database point), we bypass the graph search to reduce latency. This planning mechanism is detailed in Appendix~\ref{sec:queryPlanning}. For our experiments, we choose this threshold to be 100000.

\section{Empirical Results}


We conduct extensive experiments to evaluate our learned scoring approach for filtered approximate nearest neighbor (ANN) search.

\subsection{Datasets}
We evaluate on two diverse real world datasets, further details for each dataset are provided in Appendix~\ref{sec:dataset}.
\begin{itemize}
    \item \textbf{YFCC1M}: A subset of the Yahoo Flickr Creative Commons dataset with image embeddings and metadata based labels. 
    \item \textbf{Wikipedia}: Sentence embeddings with topic/category labels derived from article and Wikipedia categories \cite{cohere2023wikipedia}.
\end{itemize}

For each dataset, we partition the query set into training and evaluation subsets, which are randomly sampled and used for learning the weights and assessing performance, respectively. Specifically, we use 11,865 training and 5,718 evaluation queries for YFCC1M, and 308 training and 464 evaluation queries for Wikipedia.

\subsection{Evaluation Metrics}
We measure \textbf{Recall@k} under different distance functions and indexing strategies. Recall@k is defined as the fraction of ground truth neighbors, computed by retrieving the k nearest vectors that satisfy the predicate constraint, that are present in the top-k retrieved results. In all our evaluations we use Recall@10.



\subsection{Comparative Evaluation}


We compare our proposed method against two commonly used baselines to evaluate the effectiveness of incorporating filter aware scoring in ANN search.

\begin{itemize}
    \item \textbf{Integrated Learning (Ours)}: Both index construction and search use the learned distance function that jointly models vector similarity and filter mismatch.

    \item \textbf{Fixed Penalty Search}: The index is constructed using standard vector distances. At query time, a manually tuned fixed penalty scoring function is used to combine distance and filter mismatch \cite{multifilterann2025}.

    \item \textbf{Post Filtering}: The index is built and searched using pure vector distances, completely ignoring filter constraints during traversal. After the search has converged, the retrieved candidates are filtered to retain only those that satisfy the filter constraints.

\end{itemize}

These results demonstrate the effectiveness of our integrated learning approach, showing that the consistent use of the learned scoring function in both indexing and search improves recall and latency. We report the average number of vector distance comparisons during search as a proxy for search latency, since it is independent of hardware and of code level optimizations.

\begin{figure}[H]
\vskip 0.2in
\begin{center}
\centerline{\includegraphics[width=\columnwidth]{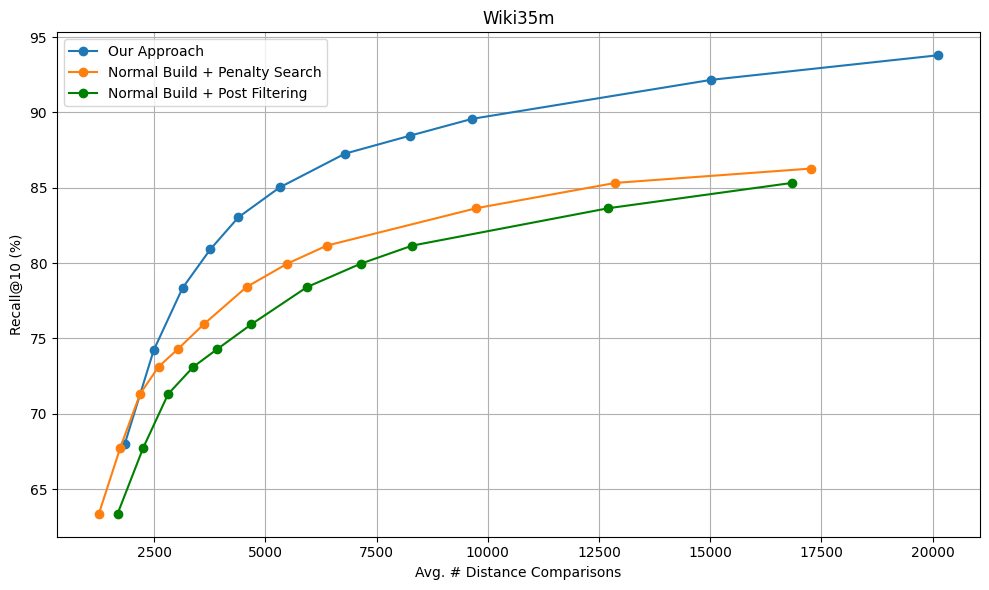}}
\caption{
Recall vs. Average Distance Comparison on the Wikipedia-35M dataset. The learned weight $w_m = 0.204148$, and 464 queries proceeded to the graph search.
}

\end{center}
\vskip -0.2in
\end{figure}

\begin{figure}[ht]
\vskip 0.2in
\begin{center}
\centerline{\includegraphics[width=\columnwidth]{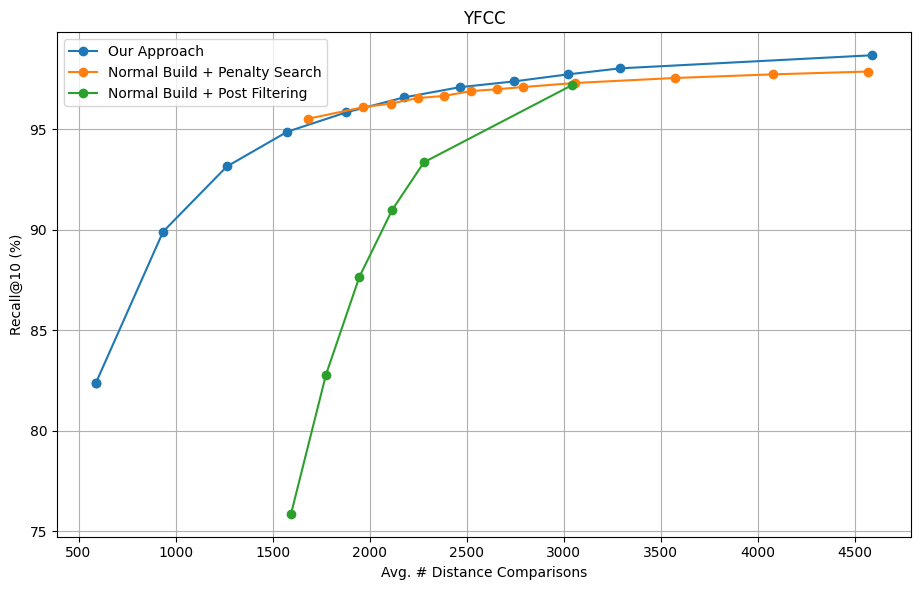}}
\caption{
Recall vs. Average Distance Comparison on the YFCC dataset. The learned weight is $w_m = 0.017787$, and 1,727 queries proceeded to the graph search.
}
\end{center}
\vskip -0.2in
\end{figure}


\begin{figure}[ht]
\vskip 0.2in
\begin{center}
\centerline{\includegraphics[width=\columnwidth]{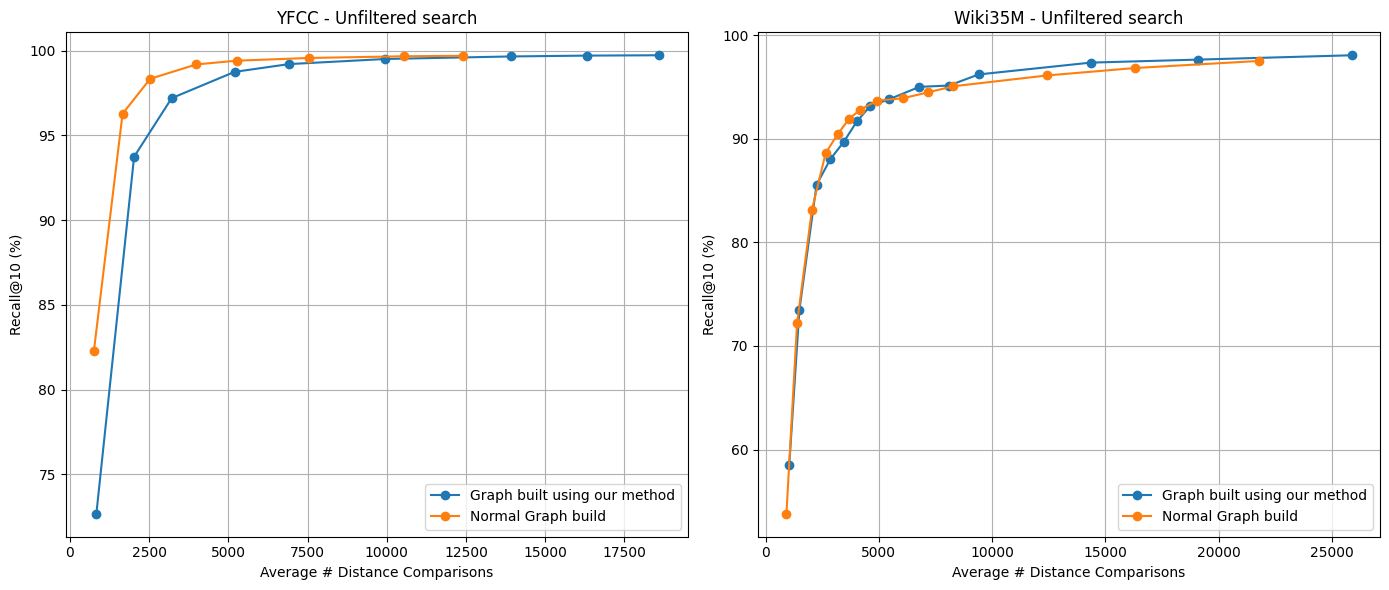}}
\caption{Unfiltered Search on different builds, this shows our index construction method preserves the quality of unfiltered search. }
\end{center}
\vskip -0.2in
\end{figure}


We also show that our index construction method preserves the quality of unfiltered search. Although the index is optimized for filtered queries, it still delivers strong performance on unfiltered queries. This shows that incorporating filter-awareness during index construction does not degrade the quality of unfiltered nearest neighbor retrieval.

\section{Discussion}

\textbf{Sensitivity to dataset statistics}:
The learned weight adapt to dataset-specific properties, including the distribution of vector distances and the sparsity of filter matches. For instance, when relevant filtered neighbors are rare, the penalty weight \( w_m \) increases to prioritize filter match. Conversely, when filter constraints are less selective, the distance plays a larger role.

\textbf{Impact of weighted distance function in index construction}:  
Incorporating the learned weighted distance function directly into index construction improves the quality of the underlying graph. By guiding edge creation to account for both vector proximity and filter similarity, the resulting graph structure clusters relevant candidates more effectively according to filter constraints. This enables more efficient searches with higher recall, as the graph encodes the trade-off directly rather than relying on heuristic edge pruning.

\section{Conclusion}
We proposed a principled, data-driven method to learn the trade-off between vector distance and filter match in filtered approximate nearest neighbor search. By formulating this as a constrained optimization problem, we derived distance function that adapt to the dataset and improve retrieval quality. Integrating these learned distance function into both index construction and search leads to substantial accuracy gains over traditional fixed-penalty heuristics.




\textbf{Future work.} 
Promising directions for future work include exploring nonlinear distance functions to better model complex interactions between vector similarity and filter compliance; investigating more expressive filter similarity metrics beyond binary matches or Jaccard distance to capture richer label relationships; and extending the framework to support more general constraints, such as continuous or hierarchical ones. Additionally, future approaches could eliminate the dependency on queries and ground-truth neighbors by learning the distance function directly from dataset-level statistics. This shift would simplify the training process and enhance generalization, as the learned function would reflect the inherent structure of the data rather than being tailored to a specific query set or labeled ground truth.

\section{Acknowledgments}
We sincerely thank Kiran Shiragur for his valuable insights and guidance in formulating the Linear Programming problem.


\bibliography{example_paper}
\bibliographystyle{icml2025}

\newpage
\appendix
\onecolumn


\begin{algorithm}[t]
\caption{FilteredDiskANN Indexing Algorithm}
\label{alg:build}
\textbf{Data:} Database $P$ with $n$ points where the $i$-th point has coordinates $x_i$; parameters $\alpha$, $L$, $R$. \\
\textbf{Result:} Directed graph $G$ over $P$ with out-degree $\le R$
\begin{algorithmic}[1]
\STATE Initialize $G$ to an empty graph
\STATE Let $s$ denote the medoid of $P$
\STATE Let $\text{st}(f)$ denote the start node for filter label $f$ for every $f \in F$
\STATE Let $\sigma$ be a random permutation of $[n]$
\STATE Let $F_x$ be the label-set for every $x \in P$
\FOR{each $i \in [n]$}
    \STATE Let $S_{F_{x_{\sigma(i)}}} = \{\text{st}(f) : f \in F_{x_{\sigma(i)}}\}$
    \STATE Let $[\emptyset; \mathcal{V}_{F_{x_{\sigma(i)}}}] \gets$ \\
    \hspace{1em} \texttt{FilteredGreedySearch}$(S_{F_{x_{\sigma(i)}}}, x_{\sigma(i)}, 0, L, F_{x_{\sigma(i)}})$
    \COMMENT{Uses weighted distance function}
    \STATE $\mathcal{V} \gets \mathcal{V} \cup \mathcal{V}_{F_{x_{\sigma(i)}}}$
    \STATE Run \texttt{FilteredRobustPrune}$(\sigma(i), \mathcal{V}_{F_{x_{\sigma(i)}}}, \alpha, R)$ to update out-neighbors
    \FOR{each $j \in N_{\text{out}}(\sigma(i))$}
        \STATE Update $N_{\text{out}}(j) \gets N_{\text{out}}(j) \cup \{\sigma(i)\}$
        \IF{$|N_{\text{out}}(j)| > R$}
            \STATE Run \texttt{FilteredRobustPrune}$(j, N_{\text{out}}(j), \alpha, R)$
        \ENDIF
    \ENDFOR
\ENDFOR
\end{algorithmic}
\end{algorithm}


\begin{algorithm}[H]
\caption{Graph Search with Weighted Distance Function}
\label{alg:search}
\begin{algorithmic}[1]
\REQUIRE Query vector $q$, label set $s_q$, graph index $G$, weight $w_m$, number of results $k$
\ENSURE Top-$k$ retrieved neighbors
\STATE Initialize priority queue $\mathcal{Q} \leftarrow$ empty min-heap of size $\mathcal{k}$
\STATE Seed $\mathcal{Q}$ with entry point(s) in $G$
\WHILE{$\mathcal{Q}$ not empty}
    \STATE Pop $v$ from $\mathcal{Q}$ with minimal distance $D(q, v)$
    \IF{$v$ not visited}
        \STATE Mark $v$ as visited
        \FOR{each neighbor $u$ of $v$ in $G$}
            \STATE Compute vector distance: $d_q \leftarrow d(q, u)$
            \STATE Compute filter match: $m_q$
            \STATE Compute distance: $D(q, u) \leftarrow  d_q + w_m \cdot (1 - m_q)$
            \STATE Insert $u$ into $\mathcal{Q}$ with priority $D(q, u)$
        \ENDFOR
    \ENDIF
\ENDWHILE
\STATE \textbf{return} Top-$k$ elements of $\mathcal{Q}$ 
\end{algorithmic}
\end{algorithm}

\section{Query Planning}
\label{sec:queryPlanning}

While the main contribution of this work is in improving graph-based indices for handling filter predicates, it is important to note that nearly all such algorithms can struggle when the query predicate is highly selective. In extreme cases where only a very small number of database points satisfy the predicate, it is often more effective to identify these points directly—such as by intersecting inverted indices for each of the query labels—and then perform a brute-force distance computation to retrieve the top-$k$ nearest neighbors. In our current implementation, we also observe certain query predicates with low, but not extremely low, selectivity for which this brute-force approach continues to be effective.

Bringing all these considerations together, our final empirical strategy is as follows: For a given query $q$ with label set $S_q$, we first estimate the number of database points likely to satisfy $S_q$ using a sample dataset and precomputed label-wise inverted indices. Based on this estimate:

\begin{enumerate}
\item If the estimated number is very small (e.g., fewer than 100,000 points), we perform a brute-force search over the satisfying subset.

\item If the estimated number is large, we run our greedy search with weighted distance function over the graph index.
\end{enumerate}

\section{Datasets}
\label{sec:dataset}
\subsection{YFCC}
 We use a 1M subset of the YFCC dataset released as part of the BigANN Filter competition. The base vectors are 192-dimensional CLIP embeddings of images, while the queries are embeddings of text descriptions. Metadata such as camera model, year, location are used to generate label-based filter predicates. Queries are constructed with either a single-label predicate or a conjunction (AND) of two labels.
 


\subsection{Wikipedia}


We present a dataset tailored for AND-query search, derived from the dataset \cite{cohere2023wikipedia}. It consists of approximately 35 million passages extracted from Wikipedia articles. Each passage is accompanied by a dense 768-dimension embedding. To construct a comprehensive label pool for each document, we combine unsupervised keyword extraction, semantic similarity, and structured metadata from Wikipedia.

\paragraph{1. Keyword Extraction}  
We employ YAKE to extract the top-$k$ keywords from each document, based on statistical properties such as term frequency and co-occurrence.  
\textit{Parameters}: \texttt{language=en}, \texttt{n-grams=1--2}, \texttt{top=100}.

\paragraph{2. Semantic Label Matching}  
We use the Universal Sentence Encoder (USE) to embed both the document and candidate labels into a shared vector space. Cosine similarity is used to select labels that are semantically close to the document, subject to a similarity threshold (0.5) and min/max label bounds(10 and 20 respectively).

\begin{figure}[ht]
\vskip 0.2in
\begin{center}
\centerline{\includegraphics[width=\columnwidth]{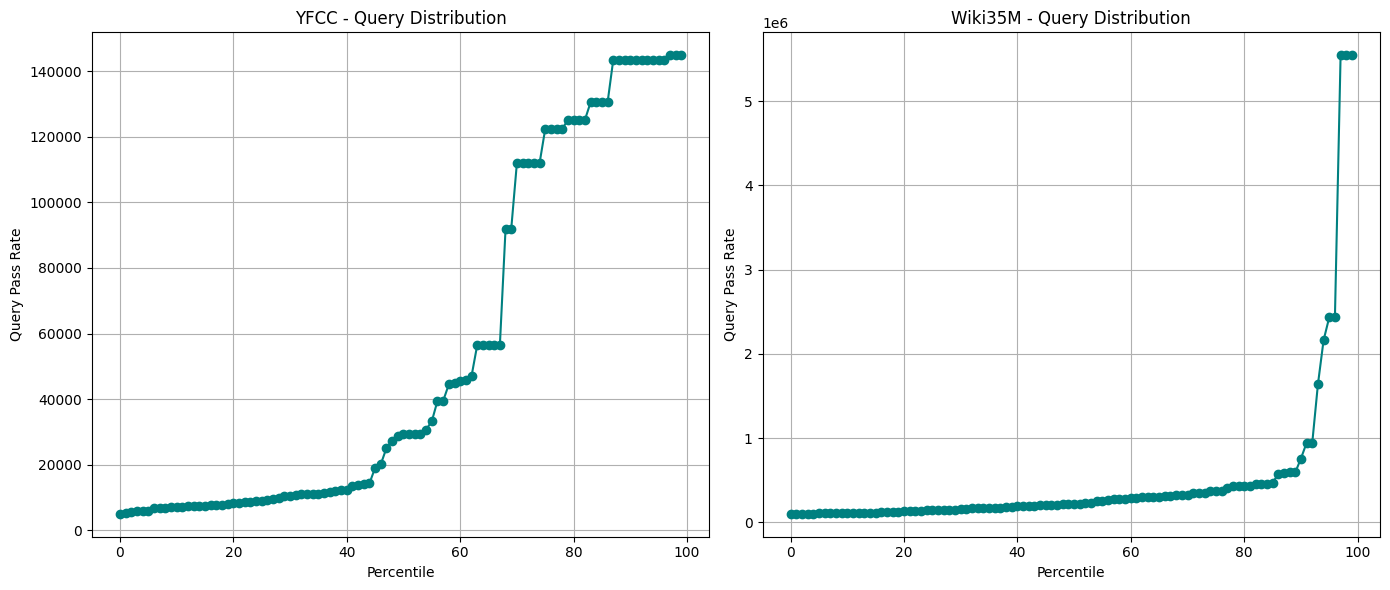}}
\caption{
Percentile refers to the ranking of queries based on their expected selectivity, with percentile 0 being the most selective (i.e., queries that pass for very few items) and percentile 99 being the least selective (i.e., queries that pass for the most items). Pass rate is defined as the total number of data points that satisfy the query conditions.
}

\end{center}
\vskip -0.2in
\end{figure}

\paragraph{3. Wikipedia Category Extraction}  
From a preprocessed Wikipedia dump (dated 2025-01-23), we extract up to 10 categories for each document title, prioritizing shorter category names (assuming they are more general). 

This hybrid approach ensures that the labels are not only diverse and context-rich but also semantically meaningful, resulting in a dataset that reflects real-world language and knowledge.

The query set is sourced from the \textit{Cohere Wikipedia Simple Embedding} dataset \cite{cohereWikiSimple}, and labels for the query vectors are generated using the same procedure described earlier. When constructing AND-query labels for search, we preferentially select the most frequent labels to ensure that a larger number of vectors satisfy each query, enabling meaningful evaluation across varying levels of selectivity.





\end{document}